# Adapting deep generative approaches for getting synthetic data with realistic marginal distributions

Kiana Farhadyar[1,2,*], Federico Bonofiglio[3], Daniela Zöller[1,2] and Harald Binder[1,2]

[1] Institute of Medical Biometry and Statistics, Faculty of Medicine and Medical Center, University of Freiburg, 79104, Freiburg, Germany

[2] Freiburg Center for Data Analysis and Modeling, University of Freiburg, 79104, Freiburg, Germany

[3] National Research Council of Italy, ISMAR, Forte Santa Teresa, Lerici - La Spezia 19032, Italy

May 14 2021

## Abstract

Synthetic data generation is of great interest in diverse applications, such as for privacy protection. Deep generative models, such as variational autoencoders (VAEs), are a popular approach for creating such synthetic datasets from original data. Despite the success of VAEs, there are limitations when it comes to the bimodal and skewed marginal distributions. These deviate from the unimodal symmetric distributions that are encouraged by the normality assumption typically used for the latent representations in VAEs. While there are extensions that assume other distributions for the latent space, this does not generally increase flexibility for data with many different distributions. Therefore, we propose a novel method, pre-transformation variational autoencoders (PTVAEs), to specifically address bimodal and skewed data, by employing pre-transformations at the level of original variables. Two types of transformations are used to bring the data close to a normal distribution by a separate parameter optimization for each variable in a dataset. We compare the performance of our method with other state-of-the-art methods for synthetic data generation. In addition to the visual comparison, we use a utility measurement for a quantitative evaluation. The results show that the PTVAE approach can outperform others in both bimodal and skewed data generation. Furthermore, the simplicity of the approach makes it usable in combination with other extensions of VAE.

---

[*] Corresponding author: e-mail: farhadya@imbi.uni-freiburg.de

# 1 Introduction

Deep generative approaches, such as variational autoencoders (VAEs) [1] have recently received much attention. Such techniques can be used to learn the joint distribution of variables in some original data by training a specific deep neural network architecture. Subsequently, synthetic observations can be generated, which can potentially be useful for many biomedical statistical analysis settings. While many applications in the machine learning community consider image data — some of which are rather playful, such as with synthetic faces of celebrities [2]— there are several serious usage scenarios also outside image data. For example, synthetic patient data can be generated [3] to allow for data exchange or federated computing under data protection restrictions [4][5], or to complement simulation studies [6]. In addition to privacy protection or simulation studies, there are also other uses for generative models. For example, the black box nature of deep neural network approaches can be alleviated by extracting patterns from the generated data [7]. Oversampling is another application of generative models, allowing to generate more observations from a minority class before analyzing unbalanced data [8][9].

While synthetic image data may be judged according to overall visual impression, without investigating the distribution of individual pixels, classical statistical properties, such as marginal distributions, skewness and potential bimodality, should be faithfully reproduced for other types of biomedical data. We specifically investigate VAEs, which have delivered promising results in many applications but have problems with data deviating from unimodal symmetric distributions, due to Gaussian model components. Therefore, we introduce an extension based on pre-transformations of the original data, called PTVAE (for Pre-Transformation Variational Autoencoder), to leverage the potential of VAEs for more general biomedical applications.

There already are some VAE-based approaches for addressing data deviating from normal distributions, such as using different priors (e.g., [10] for multimodal distributions). Ostrovski et al. allow for even more flexibility in an autoregressive implicit quantile network approach (AIQN) [11], which retains the quantiles of the mapping of the original data to the VAE latent space, based on quantile regression. Yet, modifications of VAEs that target the distribution of the latent space might not be flexible enough when different variables in the original data exhibit different kinds of peculiarities in their distribution. This motivates our PTVAE approach, which introduces pre-transformations at the level of the original variables.

generative adversarial networks (GAN) are another popular class of generative approaches. While GANs are known to generate realistic single observations, they suffer from so-called mode collapse, which makes it difficult to reproduce distributions from the original data [12]. Naturally, there also are proposals for synthetic data outside the deep neural network community. For example, we recently introduced an approach based on Gaussian copula together with simple nondisclosive summaries [13]. The latter approach, as well as GANs, will be considered for comparison when evaluating the proposed PTVAE approach.

In Section 2, we provide a brief introduction to VAEs (also showing how they can be used for a combination of binary and continuous variables), before introducing the PTVAE approach, which combines VAEs with pre-transformations, and corresponding parameter optimization techniques. Then, we describe the evaluation of the PTVAE approach with a simulation study in Section 3, and stroke trial data in Section 4. We close with a discussion in Section 5. All our results can be reproduced at https://github.com/kianaf/PTVAE. We performed our computations in `Julia` version 1.3.0.

## 2  Methods
## 2.1 Variational Autoencoders (VAEs) for a combination of continuous and binary variables

Autoencoders are consisted of an encoder and a decoder, which both are multi-layer perceptrons. A multi-layer perceptron is a neural network with one input layer, one output layer and one or multiple hidden layers, where each layer corresponds to a linear combination of its inputs ($x_j$), weights ($w_j$) and bias ($b$) with a non-linear transformation on top ($g$) and the output of the layer is

$$a = g\left(\sum_j w_j x_j + b\right)$$

The encoder part reduces the dimensions of the input layer to a latent embedding and the decoder part tries to reconstruct the data from that. A variational autoencoder (VAE) [11] is a probabilistic versions of autoencoders, with a prior distribution for the latent representation. The posterior distribution $p(z|x)$ of the latent variables z given the observed variables x can be obtained via the Bayes' rule:

$$p(z|x) = \frac{p(z,x)}{p(x)} = \frac{p(z,x)}{\int p(z,x)dz}$$

This is computationally intractable because of the integral part even when z has a relatively low dimension. One solution is to use variational inference to approximate $p(z|x)$ by a $q(z|x)$ that has simpler structure. The optimization problem then is to minimize the Kullback-Leibler (KL) divergence:

$$D_{KL}(q||p) = E_{q(z|x)}[log(q(z|x)) - log(p(x,z))] + log(p_x(x))$$

Since $log(p_x(x))$ in equation (2) is constant and KL divergence is a positive value, we can minimize it by maximizing the so-called lower bound $E_{q(z|x)}[log(q(z|x)) - log(p(x,z))]$. In the VAE, the encoder models the $q_\varphi(z|x)$, where the parameters $\varphi$ are called the encoder weights and biases, and the decoder models $p_\theta(x|z)$, with parameters $\theta$. The lower bound loss function for optimizing $\varphi$ and $\theta$ by gradient descent then is

$$loss(x_i) = -E_{q_\varphi(z|x_i)}[log\, p_\theta(x_i|z)] + KL(q_\varphi(z|x_i)||p_\theta(z))$$

The first term on the right-hand side corresponds to reconstruction loss. The second term is the *Kullback-Leibler* divergence between the approximated posterior and the prior distribution. In the approximate posterior, the individual elements $z_i$, $i = 1, ..., l$ of latent representation $z$ each are assumed to follow a normal distribution. Therefore, the architecture of encoder is designed in a way that it encodes the data to $\mu_{z_i}$ and $\sigma_{z_i}$, and the latent $z$ is sampled from $N(\mu_{z_i}, \sigma_{z_i})$. For the prior $p_\theta(z)$ a standard normal distribution is assumed, making the KL divergence computable in a straightforward manner [1].

There are two methods for obtaining synthetic data from a trained VAE: 1) sampling z from the posterior given the original data or 2) sampling z from the standard normal distribution (prior), followed in both cases by using the obtained values of z as input for the decoder. The second better preserves privacy because the original data can influence the synthetic data only via the trained parameters.

To be able to simultaneously generate continuous as well as binary variables, we used an architecture that has separate parts corresponding to the two variable types. Specifically, the decoder contains hidden layers $f$. These serve as the joint basis for continuous and binary covariates, e.g., for representing correlation patterns between the two types of variables. The distribution parameter vectors $\mu_{\hat{x}}$, $\sigma_{\hat{x}}$ for the continuous variables and $\pi_{\hat{x}}$ for the binary variables are obtained as functions of the output of $f$. This means that values for the continuous variables are subsequently obtained by sampling from $N\left(\mu_{\hat{x},j}(f(z)), \sigma_{\hat{x},j}(f(z))\right)$ and for the binary variables by sampling from $Bernouli\left(\pi_{\hat{x},k}(f(z))\right)$. Assuming that $x_{i,j}$ is representing $j$th continuous variable of the $x_i$ and $x_{i,k}$ is representing the $k$th binary variable of $x_i$ and there are $p_c$ continuous variables and $p_b$ binary variables (with $p = p_b + p_c$ being the total number of variables), the loss function is as

$$loss(x_i) = -\sum_{k=1}^{p_b} logpdf\left(Bernouli(\pi_{\hat{x},k}(f(z)), x_{i,k}\right) - \sum_{j=1}^{p_c} logpdf\left(Normal\left(\mu_{\hat{x},j}(f(z)), \sigma_{\hat{x}}(f(z))\right), x_{i,j}\right) + KL(q_\varphi(z|x_i)||p_\theta(z))$$

## 2.2 Pre-Transformation Variational Autoencoder (PTVAE)

In our proposed approach, the pre-transformation variational autoencoder (PTVAE), the underlying idea is to transform the original data in a way that it can be better modeled by a VAE. Specifically, we aim to transform the data to a symmetric unimodal distribution, before feeding into the VAE. Then, with back-transformations of the VAE output, we would have more realistic synthetic data for our more complex distributions. The pipeline for this approach is illustrated in Figure 2. The proposed transformations are represented in Figure 3, and described in more detail in Section 2.2.1.

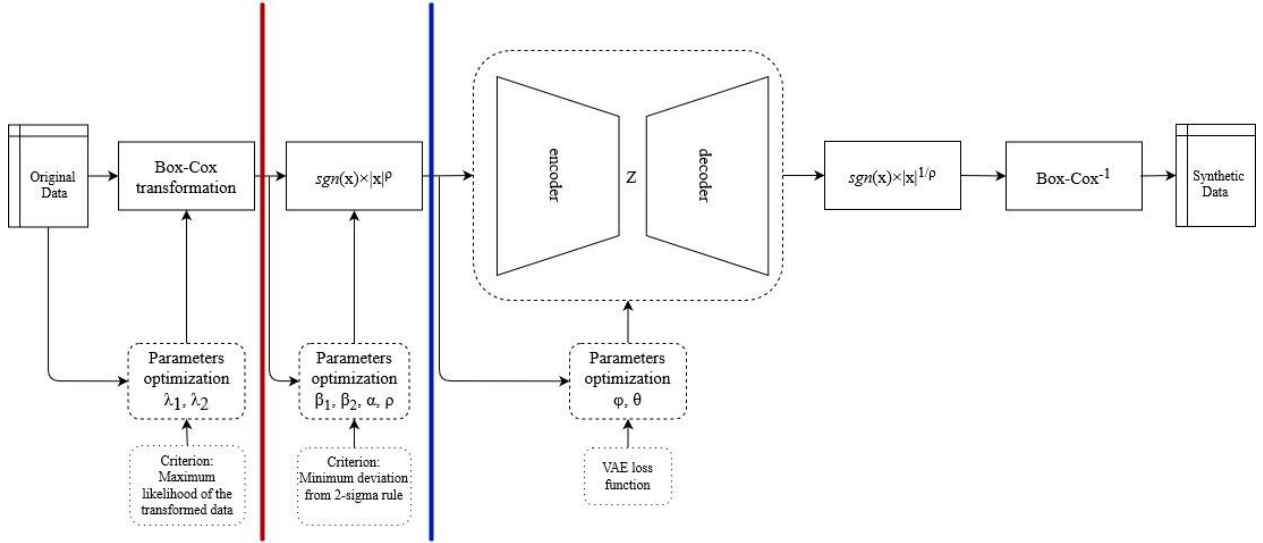

**Figure 1.** The workflow of the PTVAE approach: First the Box-Cox transformation is applied on the original data with parameters $\lambda_1$ and $\lambda_2$ optimized based on the maximum log likelihood criteria, before using the power function with parameters $(\beta_1, \beta_2, \alpha, \rho)$. By then training the VAE and applying the corresponding back-transformation functions, more realistic synthetic dataset could be obtained.

## 2.2.1 Transformations

We consider a combination of two transformations: A first transformation for removing potential skewness and then a second to move potential bimodal distributions closer to unimodal distributions.

The Box-Cox transformation [15] is a family of power transformations, which can transform the data to a symmetric distribution from a skewed distribution. Box and Cox (1964) have suggested that

$$f_{BoxCox}(y; \lambda_1, \lambda_2) = \begin{cases} \frac{(y + \lambda_2)^{\lambda_1} - 1}{\lambda_1} & \lambda_1 \neq 0 \\ \ln(y + \lambda_2) & \lambda_1 = 0 \end{cases}$$

can transform data y to an approximately normal distribution, where $\lambda_2$ is a shifting value to make the data positive and $\lambda_1$ is the main parameter of Box-Cox transformation.

The power function $x^\rho$ with the odd power ($\rho = 2k + 1, \ k = 1, 2, ..., N$) can transform bimodal data such that it becomes closer to a unimodal distribution if we can shift and scale values in a way such that the two peaks of the bimodal distribution become close to -1 and +1 and the valley becomes close to zero. Specifically, we have to find the best values for the shifting parameter $\alpha$ and scaling parameters $\beta_1$ and $\beta_2$ and also the power $\rho$. To have a differentiable function in respect to all parameters, we should change it such that $\rho$ can accept all values (not only the odd numbers). For this we use $sgn(x)|x|^\rho$ instead of $x^\rho$, in order to have the same behavior for all different values of $\rho$.

After getting the output from VAE, the data need to be transformed back. Hence, we perform

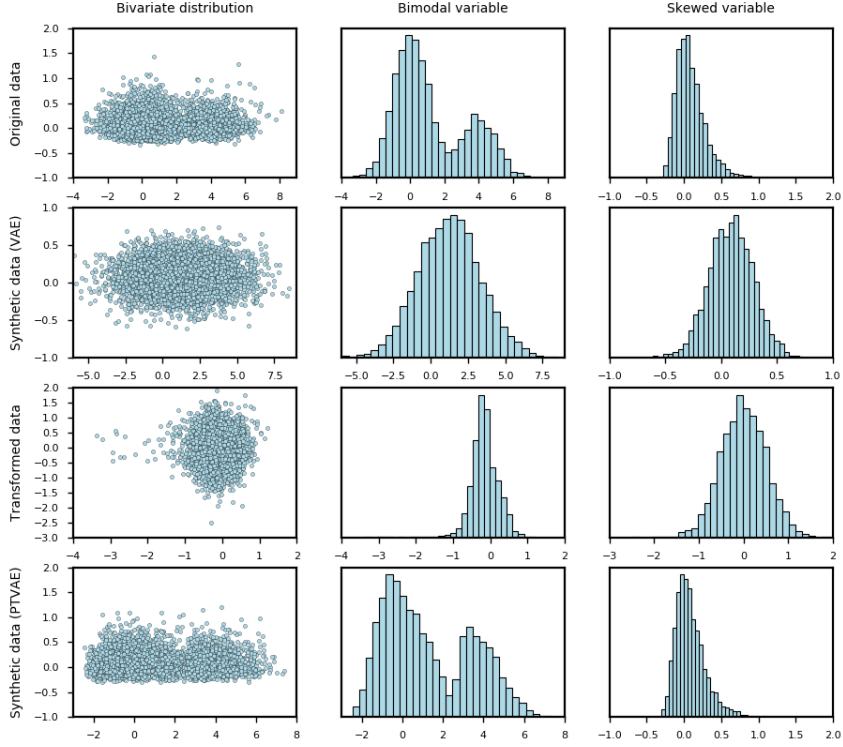

**Figure 2.** Toy example for illustrating the proposed transformations, with bivariate distributions of two variables shown in the first column, and the two marginal distributions in the second and third column. The first row shows the original data, the second row synthetic data from a standard VAE. The transformed data after Box Cox and power function transformations are shown in the third row, and synthetic data generated by the proposed PTVAE approach in the last row.

$sgn(x)|x|^{\frac{1}{\rho}}$ and shifting and scaling back on the output of VAE and then the Box-Cox transformation inverse.

$$f_{BoxCox}^{-1}(y) = \begin{cases} \sqrt[\lambda_1]{\lambda_1 y^{(\lambda)} + 1} - \lambda_2 & \lambda_1 \neq 0 \\ e^{y^{(\lambda)}} - \lambda_2 & \lambda_1 = 0 \end{cases}$$

Since this inverse has a specific domain and the VAE cannot guarantee this domain, if $\lambda_1$ was even and $\lambda_1 y^\lambda + 1 < 0$, we truncate the data at zero before back-transformation.

### 2.2.2 Parameter optimization

As seen in Figure 2, we have three parts where parameters need to be optimized. The parameters of Box-Cox transformation ($\lambda_1$ and $\lambda_2$), the parameters of $sgn(x)|x|^\rho$ function $\alpha$, $\beta_1$, $\beta_2$ and $\rho$ and the networks parameters of encoder and decoder in VAE. As $\lambda_2$ is only a shifting parameter to make the values larger than zero, it can be determined in a straightforward manner. Estimation of $\lambda_1$ for Box-Cox transformation is a real optimization problem. There are different approaches that can be used for this. Using maximum likelihood of transformed data, as suggested by Box and Cox[14], we determine $\lambda_1$ by minimizing the negative log-likelihood of a normal distribution by gradient descent. To optimize the parameters for the bimodality transformation, we need a criterion that reflects closeness to a unimodal distribution. We considered maximum likelihood and bimodality coefficient. As seen in Figure 3, where we applied the transformations to toy data, using maximum likelihood as a criterion for optimizing the second transformation did not give adequate results as the variance decreased too strongly. Unfortunately, also using the bimodality coefficient ($b = \frac{\gamma^2+1}{\kappa}$, where $\gamma$ is the skewness and $\kappa$ is the kurtosis) to judge closeness did not result in approximately normally distributed data because it decreased the bimodality by increasing the kurtosis as is shown in Figure 3. Thus, we devised a 2-sigma rule as a normality criterion to make the peaks closer and to simultaneously keep the shape of the tails close to a normal distribution as much as possible.

$$2 - sigma_{criterion}(x)$$
$$= |quantile(0.84, x) - Median - \sigma_x| + |Median - quantile(0.16, x) - \sigma_x|$$
$$+ |quantile(0.16, x) - quantile(0.02, x) - \sigma_x|$$
$$+ |quantile(0.98, x) - quantile(0.84, x) - \sigma_x|$$

We therefore optimize the parameters $\beta_1, \beta_2, \alpha$ and $\rho$ in a way such that all the differences between median and 84$^{th}$ percentile, median and 16$^{th}$ percentile, 2$^{nd}$ and 16$^{th}$ percentile and 84$^{th}$ and 98$^{th}$ percentile become close to σ. For that, we need to minimize $2 - sigma_{criterion}(x)$. In this optimization problem we try to find the local minimum of this criterion using the gradient concept. The gradient of a function is a way to find the partial derivative of that function with respect to a specific variable in a given point. Taking the gradients in each step of the optimization guides the optimizer on how to change the variable in the next step for getting closer to the local minimum. Algorithm 2 shows the algorithm of parameter estimation for $sgn(x)|x|^\rho$ transformation.

---

Algorithm 2. Pseudo-code description of the $\alpha$, $\beta_1$, $\beta_2$ and $\rho$ optimization

---

x_BC: output array of Box-Cox transformation for a specific variable
$\alpha = 0$
$\beta_1, \beta_2 = -1, 1$
$\rho = 1$
for i = 1 to I do
  for j = 1 to *epochs* do
    G = gradient($\alpha \rightarrow 2sigma_{metric}(x\_BC)$
    Update($\alpha, G$)
  end for
  for j = 1 to *epochs* do
    G = gradient($\rho \rightarrow 2sigma_{metric}(x\_BC)$
    Update($\rho, G$)
  end for
  for j = 1 to *epochs* do
    G = gradient($\beta_1 \rightarrow 2sigma_{metric}(x\_BC)$
    Update($\beta_1, G$)
  end for
  for j = 1 to *epochs* do
    G = gradient($\beta_2 \rightarrow 2sigma_{metric}(x\_BC)$
    Update($\beta_2, G$)
  end for
end for

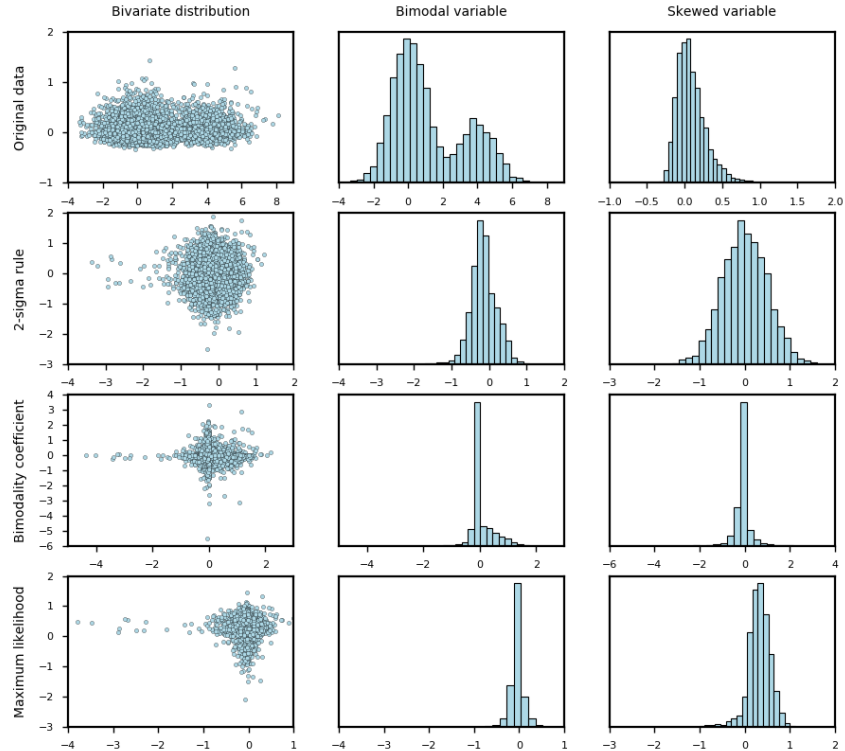

**Figure 3.** Toy example for illustrating the differences between different criteria for bimodality transformation. The first row shows the bivariate distribution comprising of a bimodal and a skewed variable. The other rows show the transformed data after applying a Box Cox transformation and also the power transformation, but with different optimization criteria. The second row shows the results for 2-sigma rule criterion. The third row is presenting the results when using the bimodality coefficient and the last one corresponds to maximum likelihood estimation.

The parameters of VAE are the weights and biases of the encoder and decoder ($\varphi, \theta$). These are obtained based on the loss function by gradient descent. It should be mentioned that for speeding the optimizations up, we shift the data by subtracting the mean and scale it with dividing by two standard deviations before each optimization step [15]. This needs to be considered in the backward process as well.

## 3 Evaluation with simulated data
## 3.1 Simulation design

For the evaluation of our method, we use a published simulation design that is based on a large breast cancer study[16][17]. Specifically, we used the specification published on Zenodo [18]. The sample size is equal to 2500, and there are 21 variables, where 12 are binary and the rest continuous variables. None of these continuous variables had a bimodal distribution. So, we extended it to include a bimodal distribution. For this, we selected one of the binary variables representing whether a patient was exposed to a specific treatment or not and we produced a bimodal distribution based on the value of this variable (we sampled from $N(0,1)$ if $x = 0$ and sampled from $N(4,1)$ if $x = 1$). The resulting bimodal distribution also is attractive for evaluating our proposed approach, because this distribution is not symmetric due to the different frequency of 0 and 1. On the other hand the modes are not very far and they have overlaps, which it makes that harder for the VAE to imitate the data. Moreover, in this dataset, we have slightly and severely skewed variables and this can evaluate the performance of the PTVAE for skewed distributions.

As the first step, we optimized the transformations' parameters. After the parameters optimization for the pre-transformations of our method, we trained the VAE with the mini-batch size of 50 and learning rate equal to 0.01 for 100 epochs. In the network of our encoder, we used 21 inputs (the number of variables) and a hidden layer of 21 nodes. Since we used a three-dimension latent space, the number of nodes for $\mu$ and $\sigma$ is three. In the decoder network, again we have a 21-node hidden layer and we use nine (number of continuous variables) nodes for $\hat{\mu}$ and $\hat{\sigma}$ and 12 (number of binary variables) nodes for node for $\hat{\pi}$. Then, we can get the data, sampling from $N(\hat{\mu}, \hat{\sigma})$ for continuous variables and sampling from $Bernouli(\hat{\pi})$ for binary variables. At the end, we rounded the values of variables, which had only integer values in the original data to the integer values to obtain realistic data. We used the *tanh* activation function for the hidden layers.

## 3.2 Approaches for comparison

For evaluation of our method, we compared the results of PTVAE on the simulation design data, with the standard VAE, generative adversarial networks (GAN)[12], which is another popular class of deep generative approaches, the VAE with an autoregressive implicit quantile network approach (AIQN) [11] (called QVAE in the following), and an approach based on Gaussian copula based approach, which we have recently proposed, with the first four moments (Norta-J) [13]. The standard VAE had the same network architecture as the PTVAE approach, which is described in 3.1., and we trained it for 100 epochs with mini-batch size of 50 and learning rate of 0.01.

In the QVAE approach, quantile regression is used to allow for more flexibility in the VAE latent space. Specifically, a neural network, embedded in the latent space, implements the quantile regression for different dimensions. In the quantile network, we used a random number $0.05 < \tau < 0.95$ as an input of quantile network and we produced a new latent z as an input of the decoder. For the quantile network architecture, because we have a conditional network based on previous dimensions and $\tau$, we need to use the best order of z dimensions. For that, we used the Kolmogrov-Smirnov test, to find out which order makes the conditional distribution closer to a normal distribution. For each dimension $i$ we have the value $\tau$ and $z_1, z_2, ..., z_{i-1}$ as the inputs of the network, a hidden layer (3 nodes) and an output layer (one node). Training the networks with quantile regression loss function for each dimension, we would have new latent value for that specific $\tau$. (For more information on the details of the QVAE approach see [11]). The learning rate of the quantile network was 0.01 and we trained the quantile network for 50 epochs. The VAE part of the QVAE had the same network architecture as standard VAE and PTVAE.

GANs comprise two multiple layer perceptrons, one called the discriminator and the other the generator. The generator part is responsible for generating synthetic data and at the same time the discriminator should be able to distinguish between real data and generated data. The better the generator, the harder it is to distinguish for the discriminator. After training the generator to fool the discriminator, which is also trained, the generator should be able to generate realistic synthetic data. We used a generator with 15 random inputs, a hidden layer of 30 nodes and 21 outputs. The activation function of the hidden layer was leaky relu and the output has a sigmoid activation function for binary values and no activation function for continuous ones. The discriminator had 21 inputs, a 25-node hidden layer and a node for the output. The hidden layer had again leaky relu activation functions and the output layer a sigmoid activation function. The learning rate was 0.0001. The training mini-batch size was 100 and we trained generator and discriminator once in each epoch for 400 epochs.

### 3.3 Evaluation criteria

For evaluation, we use visual comparisons and a utility measure based on propensity scores. The propensity score based measure was first introduced by Karr et al. [19] and extended by Snoke et al. [20] to be used as a utility measure for synthetic data. The propensity score is the conditional probability of a group membership given specific covariates. If we can show the similarity of the probability for being a member of synthetic and the probability of original data, we can claim that these two datasets have similar distributions. To do that we merge the original and synthetic data with the label 1 for synthetic ones and 0 for original ones and fit a classification and regression tree (CART) model on this data. Then we can predict the $\hat{p}_i$ for $x_{i=1,2,3,...,N}$ ($N = n_{syn} + n_{Orig}$), which is the probability of each data record being a member of synthetic data. The more differentiation $\hat{p}_i$ has from the ratio of synthetic data size to the merged data size ($c = \frac{n_{syn}}{N}$), the less similar are original and synthetic data. Hence, propensity score utility or propensity score mean-squared error (pMSE) can be measured as below:

$$pMSE = \frac{1}{N} \sum_{i=1}^{n} (\hat{p}_i - c)^2$$

Snoke et al [22] also suggested the pMSE-ratio (the ratio of pMSE calculated in pMSE to its estimated null expectation) and for the CART method they suggested a permutation resampling approach for estimating null standard deviation and null expectation. For this estimation, we permute the labels of the merged data and compute the pMSE for each permutation. Then the $\overline{pMSE_{perm_j}}$ would be the null expectation and the null standard deviation is equal to $sd(pMSE_{perm_j})$ when $j$ is the permutation index. Therefore, the pMSE-ratio is $\frac{pMSE}{\overline{pMSE_{perm_j}}}$. In this criterion, the smaller values indicate the better synthetic data. For the CART approach, we used a minimum leaf size of 20 and the maximum depth of 25.

### 3.4 Results

Table 1 shows the results for the pMSE based criteria. The proposed PTVAE approach has the best performance based on both of these criteria, followed by the Norta-J approach. Figure 5, shows exemplary synthetic data for three continuous variables (slightly skewed, severely skewed, and bimodal). In the slightly skewed variable, we can see that PTVAE is generating the skewness and the range of data better than the other methods, again followed by Norta-J, the only difference being the range of data. The GAN approach is generating inadequate symmetric distributions. The standard VAE and QVAE approaches are generating distributions closer to a normal distribution with less skewness. The QVAE is moving the peak and the standard VAE is generating some negative values and shorter tails. For the severely skewed variable, PTVAE and Norta-J are preserving the skewness. Again, the GAN is generating a symmetric distribution, and standard VAE and QVAE are generating negative values to compensate the skewness. In the bimodal variable, PTVAE is the only approach that generates bimodality. Norta-J is producing a skewed data, and VAE and GAN approaches are generating unimodal symmetric distributions. The QVAE approach performs somewhat better than standard VAE for bimodality.

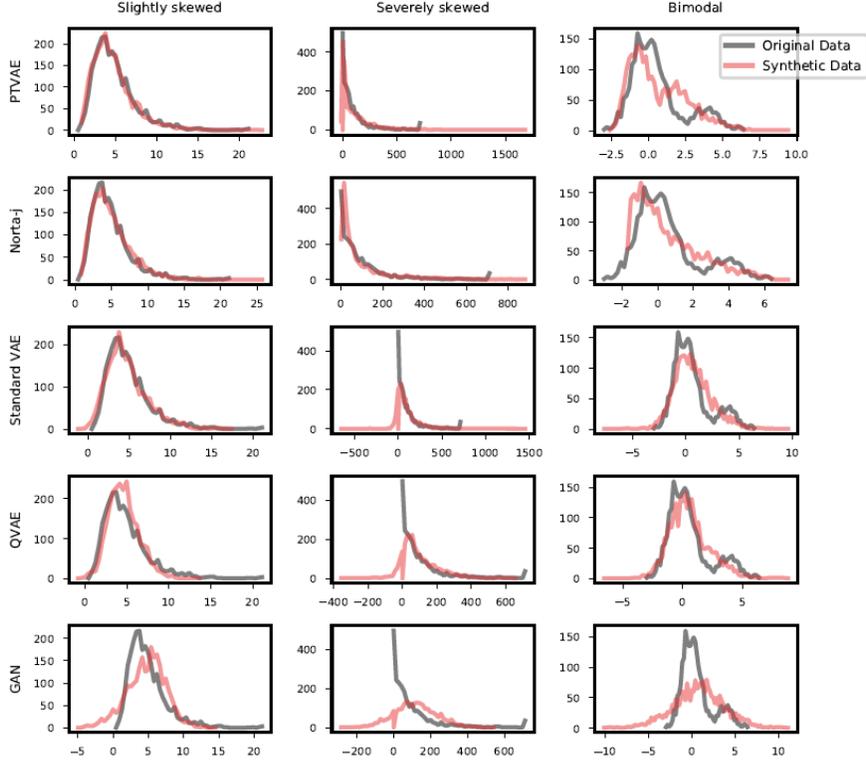

**Figure 4.** Densities of original simulated data (grey), and synthetic data (red) for three exemplary variables

**Table 1.** Comparison of synthetic data generated from original simulated data, evaluated by two pMSE based criteria.

| Measure | Methods | | | | |
| --- | --- | --- | --- | --- | --- |
| | **PTVAE** | **VAE** | **QVAE** | **GAN** | **Norta-J** |
| pMSE | 0.09 | 0.10 | 0.13 | 0.18 | 0.10 |
| pMSE_ratio | 1.71 | 1.93 | 2.46 | 3.45 | 1.83 |

## 4 Real data application

For a comparison on real data, we considered the IST dataset, which originates from a large international multicenter clinical trial stroke patients [22]. Specifically, we used a subset that was also considered in [10] for investigating the Gaussian copula approach, .i.e., 15804 patients from European centers. Moreover, according to mentioned study, we just considered and eight variables: Balanced randomized allocation of aspirin (RXASP= yes/no), heparin (RXHEP= yes/no), systolic blood pressure (RSBP), atrial fibrillation (RATRIAL= yes/no), level of consciousness (RCONSC= drowsy, unconscious or alert), gender (SEX= male vs female), age and death occurrence (FDEAD= yes/no). Among these features, Blood pressure and age are continuous, the level of consciousness is categorical (with three different values), and the rest are binary. The variable age has a potential slight bimodality. We change the level of consciousness to two binary variables (RCONSC1 = drowsy/alert and RCONSC2= unconscious/ alert) as in [10]. We removed rows with missing values, resulting in 15742 records. In PTVAE and VAE, the encoder has nine inputs (the number of variables) and a hidden layer of nine nodes. Like in simulated data, we used a three-dimensional latent space. Therefore, the number of nodes for $\mu$ and $\sigma$ is three as well. In the decoder network, we have a nine-node hidden layer and we use two (number of continuous variables) nodes for $\hat{\mu}$ and $\hat{\sigma}$ and seven (number of binary variables) nodes for node for $\hat{\pi}$. Then, we can get the data, sampling from $N(\hat{\mu},\hat{\sigma})$ for continuous variables and sampling from $Bernouli(\hat{\pi})$ for binary variables. At the end, we rounded the values of variables, which had only integer values in original data to the integer values to obtain realistic data. We used the *tanh* activation function for the hidden layers. We trained these two networks with the mini-batch size equal to 200 and learning rate of 0.1 for 100 epochs.

In the GAN approach, we tuned the parameters and used a generator with 15 inputs, a hidden layer of 30 nodes and nine outputs. The activation function of the hidden layer was leaky relu and the output has the sigmoid activation function for binary values (with rounding that made it binary) and no activation function for continuous ones. The discriminator had 9 inputs, 25 hidden layer and a node for the output and the hidden layer had again leaky relu activation function and the output layer the sigmoid activation. The learning rate was 0.00001 and mini-batch size of 100 and we trained generator and discriminator once in each epoch for 400 epochs.

The decision trees were fitted with the minimum leaf size of 20 and the maximum depth of 25. Table 2 shows the performance in terms of the pMSE based criteria. Similar to the simulated data, the PTVAE approach has the best performance. A visual comparison is also given in Figure 8. The PTVAE is the only method that can produce the potential bimodality. The Norta-J approach and the VAE approaches have similar performance.

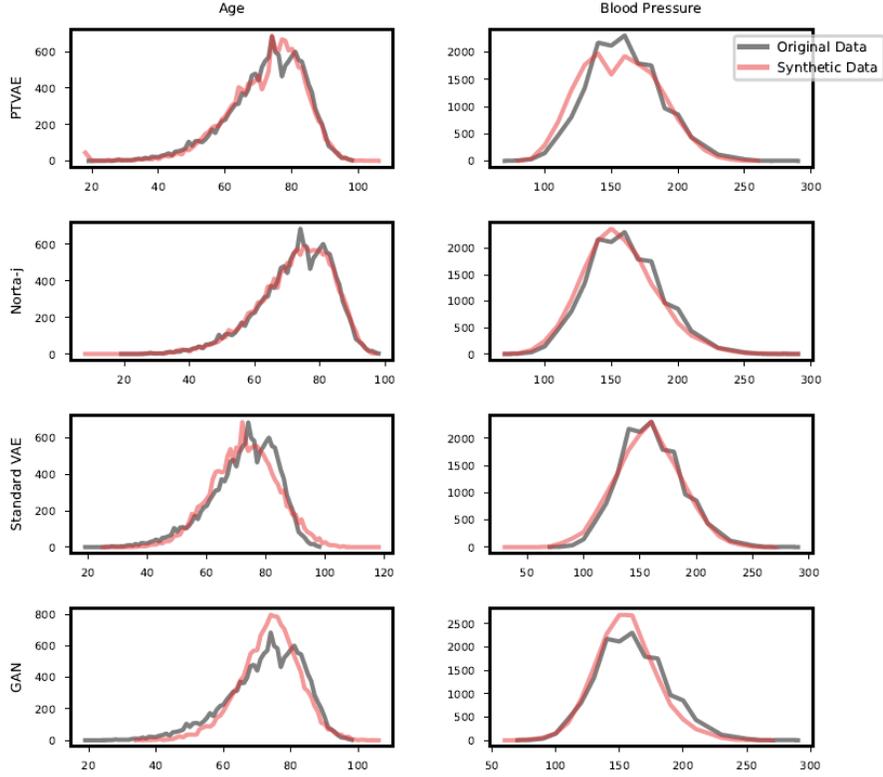

**Figure 5.** Plots of two continuous variables of IST dataset for different methods

**Table 2.** pMSE based criteria for the comparison of synthetic data generation for the IST data

| Measure | Methods | | | |
| --- | --- | --- | --- | --- |
| | **PTVAE** | **VAE** | **GAN** | **Norta-J** |
| pMSE | 0.15 | 0.16 | 0.16 | 0.15 |
| pMSE_ratio | 12.05 | 12.46 | 14.78 | 12.75 |

# 5  Discussion

For addressing the challenge of realistic synthetic data with skewed and bimodal distributions, we proposed the PTVAE approach that adapts VAEs by two pre-transformations to handle skewed and bimodal data. We investigated the performance with a comparison to synthetic data generated by other deep generative approaches, namely standard VAEs, QVAEs, and GANs. In addition to those methods, we compared our method with Norta-J as a simpler statistical approach. We used a utility measurement based on the CART method for evaluation, and according to this criterion, we could show that proposed PTVAE can generate more realistic synthetic data. In the case of bimodality in the data, PTVAE was the only method that could generate bimodal distributions. Moreover, PTVAE can generate realistic data for skewed variables. This is an important improvement over VAEs. Therefore, we are confident that the

PTVAE approach is more generally a good complement for the family of VAEs, and its simplicity could be used in combination with many extensions of VAEs. Thus, it becomes feasible to generate high-quality synthetic clinical data, e.g., for research under data protection constraints.

## Acknowledgements

This work has been supported by the Federal Ministry of Education and Research (BMBF) in Germany in the project MIRACUM (FKZ 01ZZ1801B).